\documentclass[preprint]{elsarticle}
\usepackage[ruled,vlined]{algorithm2e}
\usepackage{amsmath}
\usepackage{bm}
\usepackage{booktabs}
\usepackage[margin=1in]{geometry}
\usepackage{graphicx}
\usepackage{lineno}
\usepackage{tabularx}
\usepackage{xcolor}
\usepackage[colorlinks=true]{hyperref}

\modulolinenumbers[5]
\graphicspath{{figs/}}

\journal{Journal of \LaTeX\ Templates}

\begin{document}


\begin{frontmatter}

\title{Gradient-enhanced physics-informed neural networks for forward and inverse PDE problems}

\author[1]{Jeremy Yu}
\author[2]{Lu Lu\corref{mycorrespondingauthor}}
\author[3]{Xuhui Meng}
\author[3,4]{George Em Karniadakis}
\cortext[mycorrespondingauthor]{Corresponding author. Email: lulu1@seas.upenn.edu}

\address[1]{St. Mark's School of Texas, Dallas, TX 75230, USA}
\address[2]{Department of Chemical and Biomolecular Engineering, University of Pennsylvania, Philadelphia, PA 19104, USA}
\address[3]{Division of Applied Mathematics, Brown University, Providence, RI 02912, USA}
\address[4]{School of Engineering, Brown University, Providence, RI 02912, USA}

\begin{abstract}
Deep learning has been shown to be an effective tool in solving partial differential equations (PDEs) through physics-informed neural networks (PINNs). PINNs embed the PDE residual into the loss function of the neural network, and have been successfully employed to solve diverse forward and inverse PDE problems. However, one disadvantage of the first generation of PINNs is that they usually have limited accuracy even with many training points. Here, we propose a new method, gradient-enhanced physics-informed neural networks (gPINNs), for improving the accuracy and training efficiency of PINNs. gPINNs leverage gradient information of the PDE residual and embed the gradient into the loss function. We tested gPINNs extensively and demonstrated the effectiveness of gPINNs in both forward and inverse PDE problems. Our numerical results show that gPINN performs better than PINN with fewer training points. Furthermore, we combined gPINN with the method of residual-based adaptive refinement (RAR), a method for improving the distribution of training points adaptively during training, to further improve the performance of gPINN, especially in PDEs with solutions that have steep gradients.
\end{abstract}


\end{frontmatter}


\section{Introduction}

Deep learning has achieved remarkable success in diverse applications; however, its use in solving partial differential equations (PDEs) has emerged only recently. As an alternative
method to traditional numerical PDE solvers, physics-informed neural networks (PINNs)~\cite{dissanayake1994neural,raissi2019physics} solve a PDE via embedding the PDE into the loss of the neural network using automatic differentiation. The PINN algorithm is mesh-free and simple, and it can be applied to different types of PDEs, including integro-differential equations~\cite{lu2021deepxde}, fractional PDEs~\cite{pang2019fpinns}, and stochastic PDEs~\cite{zhang2019quantifying,zhang2020learning}. Moreover, one main advantage of PINNs is that from the implementation point of view, PINNs solve inverse PDE problems as easily as forward problems. PINNs have been successfully employed to solve diverse problems in different fields, for example in optics~\cite{chen2020physics,lu2021physics}, fluid mechanics~\cite{raissi2020hidden}, systems biology~\cite{yazdani2020systems}, and biomedicine~\cite{sahli2020physics}.

Despite promising early results, there are still some issues in PINNs to be addressed. One open problem is how to improve the PINN accuracy and efficiency. There are a few aspects of PINNs that can be improved and have been investigated by researchers. For example, the residual points are usually randomly distributed in the domain or grid points on a lattice, and other methods of training point sampling and distribution have been proposed to achieve better accuracy when using the same number of training points, such as the residual-based adaptive refinement (RAR)~\cite{lu2021deepxde} and importance sampling~\cite{nabian2021efficient}. The standard loss function in PINNs is the mean square error, and Refs.~\cite{gu2020selectnet,mcclenny2020self} show that a properly-designed non-uniform training point weighting can improve the accuracy. In PINNs, there are multiple loss terms corresponding to the PDE and initial/boundary conditions, and it is critical to balance these different loss terms~\cite{wang2020understanding,wang2020and}. Domain decomposition can be used for problems in a large domain~\cite{meng2020ppinn,jagtap2020extended,dwivedi2020physics}. Neural network architectures can also be modified to satisfy automatically and exactly the required Dirichlet boundary conditions~\cite{lagaris1998artificial,sheng2020pfnn,sukumar2021exact}, Neumann boundary conditions~\cite{mcfall2009artificial,beidokhti2009solving}, Robin boundary conditions~\cite{lagari2020systematic}, periodic boundary conditions~\cite{dong2020method,lu2021physics}, and interface conditions~\cite{lagari2020systematic}. In addition, if some features of the PDE solutions are known a-priori, it is also possible to encode them in network architectures, for example, multi-scale and high-frequency features \cite{cai2020phase,wang2020multi,liu2020multi,yazdani2020systems,wang2020eigenvector}. Moreover, the constraints in PINNs are usually soft constraints, and hard constraints can be imposed by using the augmented Lagrangian method~\cite{lu2021physics}.

In PINNs, we aim to train a neural network to minimize the PDE residual for each PDE, and thus we only use the PDE residual as the corresponding loss for each PDE. This idea is straightforward and used by many researchers in the area, and no attention has been paid to other types of losses for a PDE yet. However, if the PDE residual is zero, then it is clear that the gradient of the PDE residual should also be zero. The idea of using gradient information has been demonstrated to be useful in other methods such as Gaussian process regression~\cite{deng2020multifidelity}. In this work, we develop the gradient-enhanced PINN (gPINN), which uses a new type of loss functions by leveraging the gradient information of the PDE residual to improve the accuracy and training efficiency of PINNs. We also combine gPINN with the aforementioned RAR method to further improve the performance.

The paper is organized as follows. In Section~\ref{sec:methods}, after introducing the algorithm of PINN, we present the extension to gPINN and gPINN with RAR. In Section~\ref{sec:results}, we demonstrate the effectiveness of gPINN and RAR for eight different problems, including function approximation, forward problems of PDEs, and inverse PDE-based problems. We systematically compare the performance of PINN, gPINN, PINN with RAR, and gPINN with RAR. Finally, we conclude the paper in Section~\ref{sec:conclusion}.

\section{Methods}
\label{sec:methods}

We first provide a brief overview of physics-informed neural networks (PINNs) for solving forward and inverse partial differential equations (PDEs) and then present the method of gradient-enhanced PINNs (gPINNs) to improve the accuracy and training efficiency of PINNs. Next we discuss how to use the residual-based adaptive refinement (RAR) method to further improve gPINNs.

\subsection{PINNs for solving forward and inverse PDEs}

We consider the following PDE for the solution $u(\mathbf{x},t)$ parametrized by the parameters $\bm{\lambda}$ defined on a domain $\Omega$:
\begin{equation} \label{eq:pde}
    f\left(\mathbf{x} ; \frac{\partial u}{\partial x_{1}}, \ldots, \frac{\partial u}{\partial x_{d}} ; \frac{\partial^{2} u}{\partial x_{1} \partial x_{1}}, \ldots, \frac{\partial^{2} u}{\partial x_{1} \partial x_{d}} ; \ldots ; \boldsymbol{\lambda}\right)=0, \quad \mathbf{x}=(x_1, \cdots,x_d) \in \Omega,
\end{equation}
with the boundary conditions
$$\mathcal{B}(u, \mathbf{x})=0 \quad \text { on } \quad \partial \Omega.$$
We note that in PINNs the initial condition is treated in the same way as the Dirichlet boundary condition.

To solve the PDE via PINNs, we first construct a neural network $\hat{u}(\mathbf{x}; \boldsymbol{\theta})$ with the trainable parameters $\boldsymbol{\theta}$ to approximate the solution $u(x)$. We then use the constraints implied by the PDE and the boundary conditions to train the network. Specifically, we use a set of points inside the domain ($\mathcal{T}_f$) and another set of points on the boundary ($\mathcal{T}_b$). The loss function is then defined as~\cite{raissi2019physics,lu2021deepxde}
$$\mathcal{L}(\boldsymbol{\theta} ; \mathcal{T})=w_{f} \mathcal{L}_{f}\left(\boldsymbol{\theta} ; \mathcal{T}_{f}\right)+w_{b} \mathcal{L}_{b}\left(\boldsymbol{\theta} ; \mathcal{T}_{b}\right),$$
where
\begin{equation} \label{eq:loss_f}
    \mathcal{L}_{f}\left(\boldsymbol{\theta} ; \mathcal{T}_{f}\right) = \frac{1}{\left|\mathcal{T}_{f}\right|} \sum_{\mathbf{x} \in \mathcal{T}_{f}} \left|f\left(\mathbf{x};\frac{\partial \hat{u}}{\partial x_{1}}, \ldots, \frac{\partial \hat{u}}{\partial x_{d}} ; \frac{\partial^{2} \hat{u}}{\partial x_{1} \partial x_{1}}, \ldots, \frac{\partial^{2} \hat{u}}{\partial x_{1} \partial x_{d}} ; \ldots ; \boldsymbol{\lambda}\right)\right|^{2},
\end{equation}
$$\mathcal{L}_{b}\left(\boldsymbol{\theta} ; \mathcal{T}_{b}\right) = \frac{1}{\left|\mathcal{T}_{b}\right|} \sum_{\mathbf{x} \in \mathcal{T}_{b}} \left|\mathcal{B}(\hat{u}, \mathbf{x})\right|^{2}.$$
and $w_f$ and $w_b$ are the weights.

One main advantage of PINNs is that the same formulation can be used  not only for forward problems but also for inverse PDE-based problems. If
the parameter $\boldsymbol{\lambda}$ in Eq.~\eqref{eq:pde} is unknown, and instead we have some extra measurements of $u$  on the set of points $\mathcal{T}_{i}$. Then we add an additional data loss~\cite{raissi2019physics,lu2021deepxde} as
$$\mathcal{L}_{i}\left(\boldsymbol{\theta}, \boldsymbol{\lambda} ; \mathcal{T}_{i}\right)=\frac{1}{\left|\mathcal{T}_{i}\right|} \sum_{\mathbf{x} \in \mathcal{T}_{i}} |\hat{u}(\mathbf{x}) - u(\mathbf{x})|^{2}$$
to learn the unknown parameters simultaneously with the solution $u$. Our new loss function is then defined as
$$\mathcal{L}(\boldsymbol{\theta}, \boldsymbol{\lambda} ; \mathcal{T})=w_{f} \mathcal{L}_{f}\left(\boldsymbol{\theta}, \boldsymbol{\lambda} ; \mathcal{T}_{f}\right)+w_{b} \mathcal{L}_{b}\left(\boldsymbol{\theta}, \boldsymbol{\lambda} ; \mathcal{T}_{b}\right)+w_{i} \mathcal{L}_{i}\left(\boldsymbol{\theta}, \boldsymbol{\lambda} ; \mathcal{T}_{i}\right).$$

In this study, we choose the weights $w_f=w_b=w_i=1$. In some PDEs, it is possible to enforce the boundary conditions exactly and automatically by modifying the network architecture~\cite{lagaris1998artificial,pang2019fpinns,lagari2020systematic,lu2021physics}, which eliminates the loss term of boundary conditions.

\subsection{Formulation of gradient-enhanced PINNs (gPINNs)}

In PINNs, we only enforce the PDE residual $f$ to be zero; because $f(\mathbf{x})$ is zero for any $\mathbf{x}$, we know that the derivatives of $f$ are also zero. Here, we propose the gradient-enhanced PINNs to enforce the derivatives of the PDE residual to be zero as well, i.e.,
$$\nabla f(\mathbf{x}) = \left(\frac{\partial f}{\partial x_1}, \frac{\partial f}{\partial x_2}, \cdots, \frac{\partial f}{\partial x_d}\right) = \mathbf{0}, \quad \mathbf{x} \in \Omega.$$
Then the loss function of gPINNs is:
$$\mathcal{L} = w_{f} \mathcal{L}_{f} +w_{b} \mathcal{L}_{b} + w_i\mathcal{L}_i + \sum_{i=1}^d w_{g_i} \mathcal{L}_{g_i}\left(\boldsymbol{\theta} ; \mathcal{T}_{g_i}\right),$$
where the loss of the derivative with respect to $x_i$ is
\begin{equation} \label{eq:loss_g}
    \mathcal{L}_{g_i}\left(\boldsymbol{\theta} ; \mathcal{T}_{g_i}\right) = \frac{1}{\left|\mathcal{T}_{g_i}\right|} \sum_{\mathbf{x} \in \mathcal{T}_{g_i}} \left| \frac{\partial f}{\partial x_{i}} \right|^{2}.
\end{equation}
Here, $\mathcal{T}_{g_i}$ is the set of residual points for the derivative $\frac{\partial f}{\partial x_{i}}$. For example, for the Poisson's equation $\Delta u = f$ in 1D, the additional loss term is
$$\mathcal{L}_{g} = w_{g} \frac{1}{\left|\mathcal{T}_{g}\right|} \sum_{\mathbf{x} \in \mathcal{T}_{g}} \left|\frac{d^3 u}{d x^3} - \frac{d f}{d x} \right|^{2}.$$
For the Poisson's equation in 2D, there are two additional loss terms:
\begin{gather*}
    \mathcal{L}_{g_1} = w_{g_1} \frac{1}{\left|\mathcal{T}_{g_1}\right|} \sum_{\mathbf{x} \in \mathcal{T}_{g_1}} \left|\frac{\partial^3 u}{\partial x^3} + \frac{\partial^3 u}{\partial x \partial y^2} - \frac{\partial f}{\partial x} \right|^{2}, \\
    \mathcal{L}_{g_2} = w_{g_2} \frac{1}{\left|\mathcal{T}_{g_2}\right|} \sum_{\mathbf{x} \in \mathcal{T}_{g_2}} \left|\frac{\partial^3 u}{\partial x^2 \partial y} + \frac{\partial^3 u}{\partial y^3} - \frac{\partial f}{\partial y} \right|^{2}.
\end{gather*}

As we will show in our numerical examples, by enforcing the gradient of the PDE residual, gPINN improves the accuracy of the predicted solutions for $u$ and requires less training points. Moreover, gPINN improves the accuracy of the predicted solutions for $\frac{\partial u}{\partial x_i}$. Although $\mathcal{T}_f$ and $\mathcal{T}_{g_i}$ ($i=1, \cdots, d$) can be different, in this study we choose $\mathcal{T}_{g_i}$ to be the same as $\mathcal{T}_f$.

\subsection{Formulation of gPINN with residual-based adaptive refinement (RAR)}

The residual points $\mathcal{T}_f$ of PINNs are usually randomly distributed in the domain, and in Ref.~\cite{lu2021deepxde} a residual-based adaptive refinement (RAR) method is developed to improve the distribution of residual points during the training process. In RAR, we adaptively add more residual points in the locations where the PDE residual is large during the network training. Here we combine RAR and gPINN to further improve the accuracy and training efficiency (Algorithm \ref{alg:rar}).

\begin{algorithm}[htbp]
\caption{gPINN with RAR.}
\label{alg:rar}
\begin{itemize}
    \item[Step 1] Train the neural network using gPINN on the training set $\mathcal{T}$ for a certain number of iterations.
    \item[Step 2] Compute the PDE residual $\left|f\left(\mathbf{x} ; \frac{\partial \hat{u}}{\partial x_{1}}, \ldots, \frac{\partial \hat{u}}{\partial x_{d}} ; \frac{\partial^{2} \hat{u}}{\partial x_{1} \partial x_{1}}, \ldots, \frac{\partial^{2} \hat{u}}{\partial x_{1} \partial x_{d}} ; \ldots ; \boldsymbol{\lambda}\right)\right|$ at random points in the domain.
    \item[Step 3] Add $m$ new points to the training set $\mathcal{T}$ where the residual is the largest.
    \item[Step 4] Repeat Steps 1, 2, and 3 for $n$ times, or until the mean residual falls below a threshold $\mathcal{E}$.
\end{itemize}
\end{algorithm}

\section{Results}
\label{sec:results}

We will apply our proposed gPINNs and gPINNs with RAR to solve several forward and inverse PDE problems. In all examples, we use the $\tanh$ as the activation function, and the other hyperparameters for each example are listed in Table~\ref{tab:hyperparameter}.

\begin{table}[htbp]
\centering
\begin{tabular}{ c | c  c  c  c  c  c  } 
\toprule
Section & Depth & Width & Optimizer & Learning rate & \# Iterations \\ 
\midrule
3.1 & 4 & 20 & Adam & 0.001 & 10000 \\ 
3.2.1 & 4 & 20 & Adam & 0.001 & 20000  \\
3.2.2 & 4 & 20 & Adam & 0.0001 & 100000 \\
3.3.1 & 4 & 20 & Adam & 0.001 & 50000 \\
3.3.2 & 4 & 20 & Adam & 0.0001 & 200000 \\
3.4.1 & 4 & 32 & Adam + L-BFGS & 0.001 & 20000\\
3.4.2 & 5 & 64 & Adam + L-BFGS & 0.001 & 20000\\
\bottomrule
\end{tabular}
\caption{\textbf{Hyperparameters used for each problem.}}
\label{tab:hyperparameter}
\end{table}

\subsection{Function approximation via a gradient-enhanced neural network (gNN)}

We first use a pedagogical example of function approximation to demonstrate the effectiveness of adding gradient information. We consider the following function
\begin{equation*}
    u(x) = -(1.4-3x)\sin (18 x), \quad x \in [0, 1],
\end{equation*}
from the training dataset $\{(x_1, u(x_1)), (x_2, u(x_2)), \cdots, (x_n, u(x_n))\}$, where $(x_1, x_2, \cdots, x_n)$ are equispaced points in $[0,1]$. The standard loss function to train a NN is
$$\mathcal{L} = \frac{1}{n} \sum_{i=1}^n |u(x_i) - \hat{u}(x_i)|^2,$$
and we also consider the following gradient-enhanced NN with the extra loss function of the gradient as
$$\mathcal{L} = \frac{1}{n} \sum_{i=1}^n |u(x_i) - \hat{u}(x_i)|^2 + w_g \frac{1}{n} \sum_{i=1}^n |\nabla u(x_i) - \nabla \hat{u}(x_i)|^2.$$
We performed the network training using different values of the weight $w_g$, including 1, 0.1, and 0.01, and found that the accuracy of gNN is insensitive to the value of $w_g$. Hence, here we will only show the results of $w_g = 1$.

\begin{figure}[hbtp]
    \centering
    \includegraphics[width = 15cm]{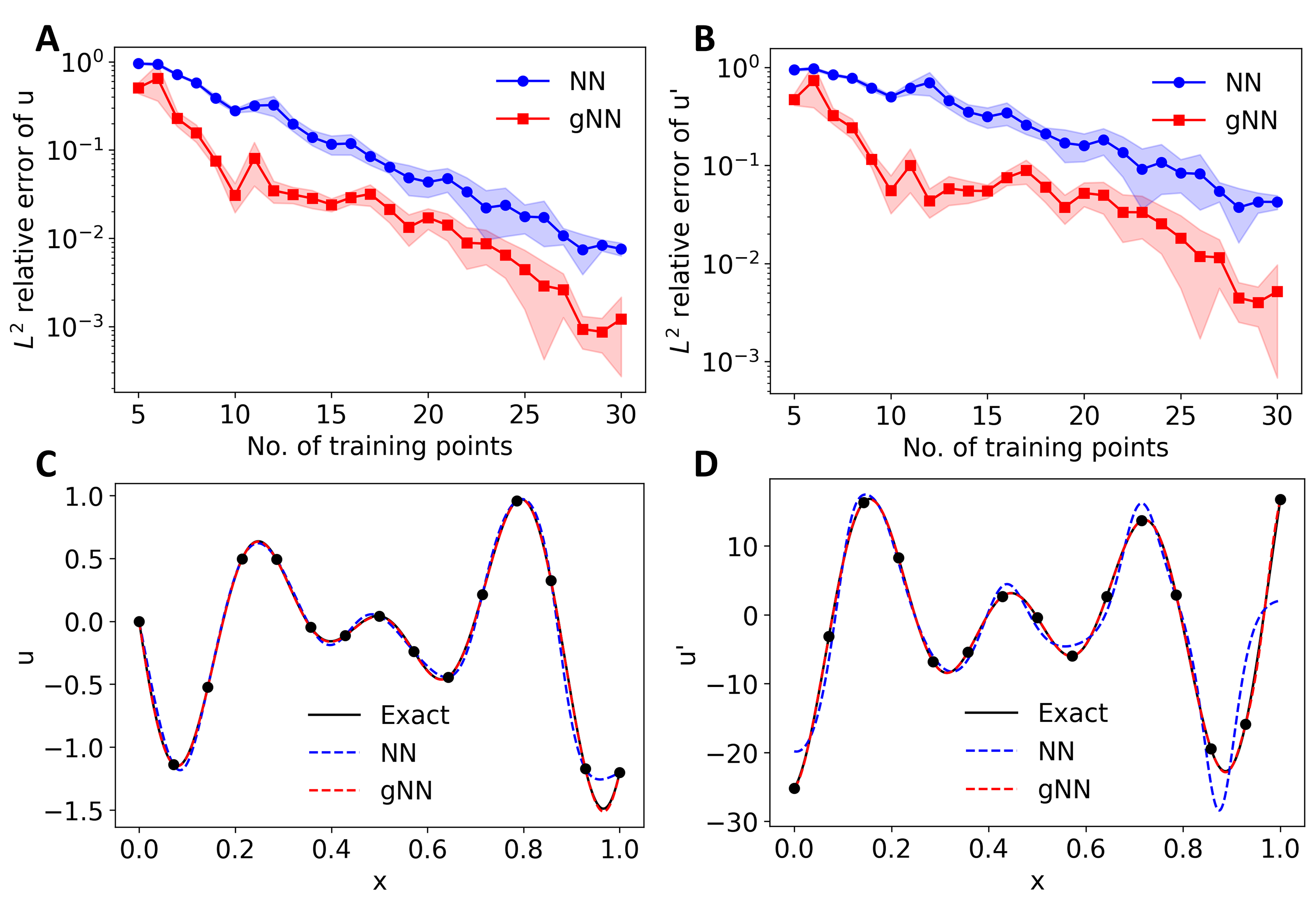}
    \caption{\textbf{Example in Section 3.1: Comparison between NN and gNN.} (\textbf{A} and \textbf{B}) $L^2$ relative error of NN and gNN for (A) $u$ and (B) $\frac{du}{dx}$ using different number of training points. The line and shaded region represent the mean and one standard deviation of 10 independent runs. (\textbf{C} and \textbf{D}) Example of the predicted (C) $u$ and (D) $\frac{du}{dx}$, respectively. The black dots show the locations of the 15 training data points.}
    \label{fig:3.1}
\end{figure}

When we use more training points, both NN and gNN have smaller $L^2$ relative error of the prediction of $u$, and gNN performs significantly better than NN with about one order of magnitude smaller error (Fig.~\ref{fig:3.1}A). In addition, gNN is more accurate than NN for the prediction of the derivative $\frac{du}{dx}$ (Fig.~\ref{fig:3.1}B). As an example, the prediction of $u$ and $\frac{du}{dx}$ from NN and gNN using 15 training data points are shown in Figs.~\ref{fig:3.1}C and D, respectively. The standard NN has more than 10\% error for $u$ and $\frac{du}{dx}$, while gNN reaches about 1\% error.

\subsection{Forward PDE problems}

After demonstrating the effectiveness of adding the gradient loss on the function approximation, we apply gPINN to solve PDEs.

\subsubsection{Poisson equation}

We first consider a 1D Poisson equation as
\begin{equation*}
    -\Delta u = \sum^4_{i = 1}i \sin(ix) + 8 \sin(8x), \quad x\in[0, \pi],
\end{equation*}
with the Dirichlet boundary conditions $u(x=0)=0$ and $u(x=\pi)=\pi$. The analytic solution is
\begin{equation*}
    u(x) = x + \sum^4_{i=1} \frac{\sin(ix)}{i} + \frac{\sin(8x)}{8}.
\end{equation*}
Instead of using a loss function $\mathcal{L}_b$ for the Dirichlet boundary conditions, we enforce it by choosing the surrogate of the solution as
$$\hat{u}(x) = x(\pi - x)\mathcal{N}(x)+x,$$
where $\mathcal{N}(x)$ is a neural network. Hence, the loss function is
$$\mathcal{L} = \mathcal{L}_f + w \mathcal{L}_g,$$
where $\mathcal{L}_f$ and $\mathcal{L}_g$ are defined in Eqs.~\eqref{eq:loss_f} and \eqref{eq:loss_g}, respectively.

\begin{figure}[hbtp]
    \centering
    \includegraphics[width=\textwidth]{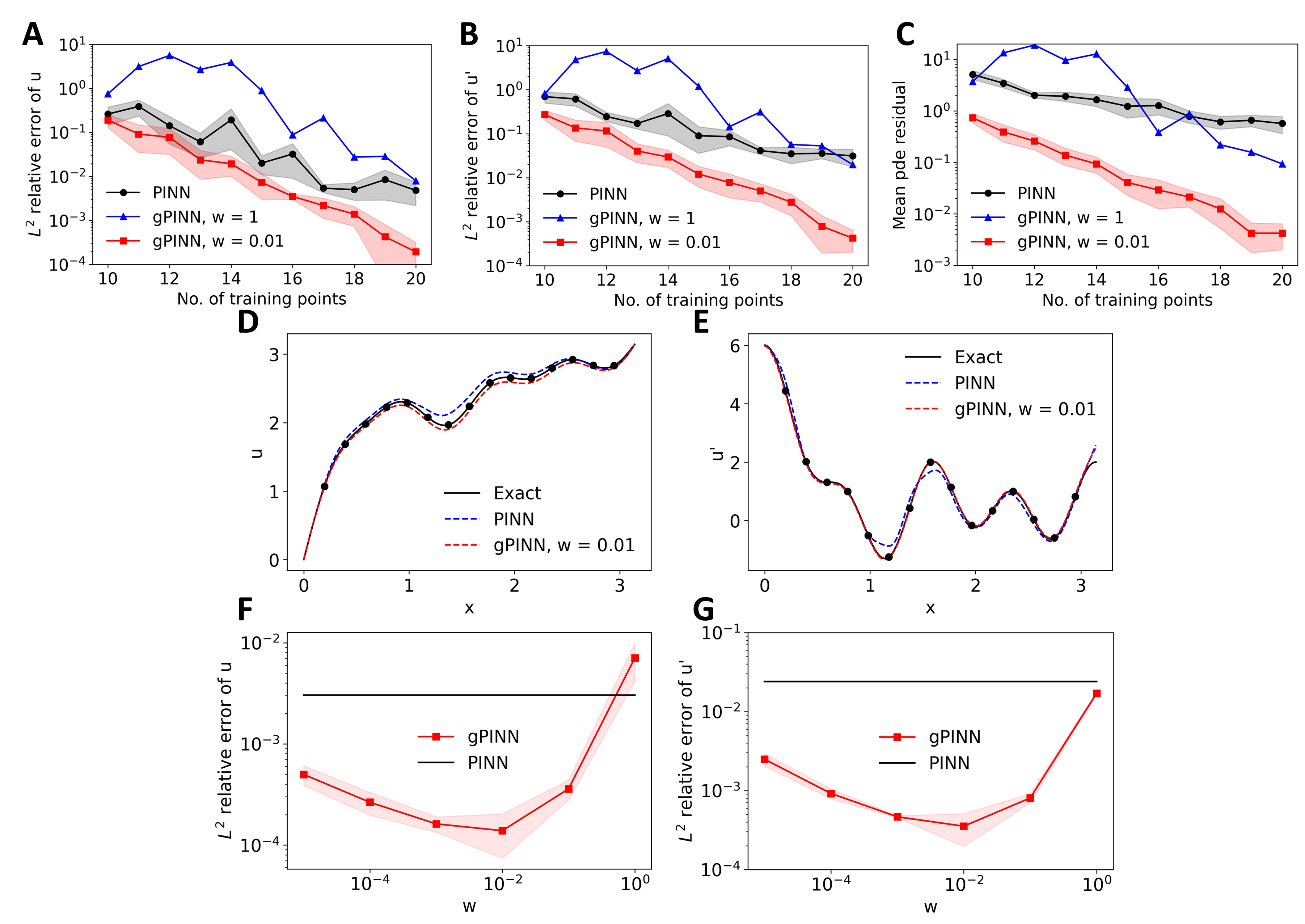}
    \caption{\textbf{Example in Section 3.2.1: Comparisons between PINN and gPINNs with the loss weight $w=1$ and $0.01$.} (\textbf{A}) $L^2$ relative error of $u$. (\textbf{B}) $L^2$ relative error of $u'$. (\textbf{C}) The mean value of the PDE residual after training. (\textbf{D} and \textbf{E}) Example of the predicted $u$ and $u'$, respectively, when using 15 training points. The black dots show the locations of the residual points for training. (\textbf{F} and \textbf{G}) $L^2$ relative errors of gPINN for $u$ and $u'$ with different values of the weight $w$ when using 20 training points. The shaded regions represent the one standard deviation of 10 random runs.}
    \label{fig:3.2.1}
\end{figure}

When increasing the number of residual points from 10 to 20, as the baseline, the $L^2$ relative error of PINN for $u$ decreases from 26\% to 0.48\% (Fig.~\ref{fig:3.2.1}A). The performance of gPINN depends on the choice of the weight $w$. For $w = 0.01$, gPINN thoroughly outperforms PINN in terms of the L$^2$ relative error of $u$ (Fig.~\ref{fig:3.2.1}A), L$^2$ relative error of $\frac{du}{dx}$ (Fig.~\ref{fig:3.2.1}B), and the mean absolute value of the PDE residual (Fig.~\ref{fig:3.2.1}C). When using 20 residual points, the $L^2$ relative error of gPINN for $u$ is about one order of magnitude smaller than PINN (Fig.~\ref{fig:3.2.1}A). An even greater improvement by gPINN can be seen in the $L^2$ relative error of $\frac{du}{dx}$ (about two orders of magnitude, Fig.~\ref{fig:3.2.1}B). gPINN outperforms PINN, because gPINN utilizes the information of the gradient and thus has a much faster convergence rate than PINN. The results of PINN and gPINN for the example of using 15 residual points are shown in Figs.~\ref{fig:3.2.1}D and E.

However, we note that if the value of $w$ is not chosen properly, gPINN may not perform well. For example, when we choose $w=1$, the accuracy of gPINN is even worse than PINN (Figs.~\ref{fig:3.2.1}A, B and C). We systematically investigated the performance of gPINN for different values of $w$ when using 20 residual points (Figs.~\ref{fig:3.2.1}F and G). When $w$ is small and close to 0, then gPINN becomes a standard PINN (the black horizontal line in Figs.~\ref{fig:3.2.1}F and G). When $w$ is very large, the error of PINN increases. There exists an optimal weight at around $w=0.01$. When $w$ is smaller than 1, gPINN always outperforms PINN.

\subsubsection{Diffusion-reaction equation}
\label{sec:diffusion-reaction}

Next we consider a time-dependent PDE of a diffusion-reaction system described as
\begin{equation*}
    \frac{\partial u}{\partial t} = D \frac{\partial^2 u}{\partial x^2} + R(x,t), \quad x \in [-\pi, \pi],~ t \in [0, 1],
\end{equation*}
where $u$ is the solute concentration, $D = 1$ represents the diffusion coefficient, and $R$ is the chemical reaction as
$$R(x,t) = e^{-t} \left[ \frac{3}{2}\sin(2x) + \frac{8}{3}\sin(3x) + \frac{15}{4}\sin(4x) + \frac{63}{8}\sin(8x) \right].$$
The initial and boundary conditions are as follows:
\begin{gather*}
    u(x, 0) = \sum^{4}_{i=1} \frac{\sin(ix)}{i} + \frac{\sin(8x)}{8}, \\
    u(-\pi, t) = u(\pi, t) = 0,
\end{gather*}
which yields the analytic solution for $u$ as
\begin{equation} \label{eq:dr_exact}
    u(x, t) = e^{-t} \left[ \sum^4_{i=1} \frac{\sin(ix)}{i} + \frac{\sin(8x)}{8} \right].
\end{equation}
Similar as the previous example, we also choose a proper surrogate of the solution to satisfy the initial and boundary conditions automatically:
$$\hat{u}(x) = (x^2-\pi^2)(1-e^{-t})\mathcal{N}(x) + u(x, 0),$$
where $\mathcal{N}(x)$ is a neural network. Here, we have two loss terms of the gradient, and the total loss function is
$$\mathcal{L} = \mathcal{L}_f + w \mathcal{L}_{g_x}+w \mathcal{L}_{g_t},$$
where $\mathcal{L}_{g_x}$ and $\mathcal{L}_{g_t}$ are the derivative losses with respect to $x$ and $t$, respectively.

\begin{figure}[htbp]
    \centering
    \includegraphics[width=\textwidth]{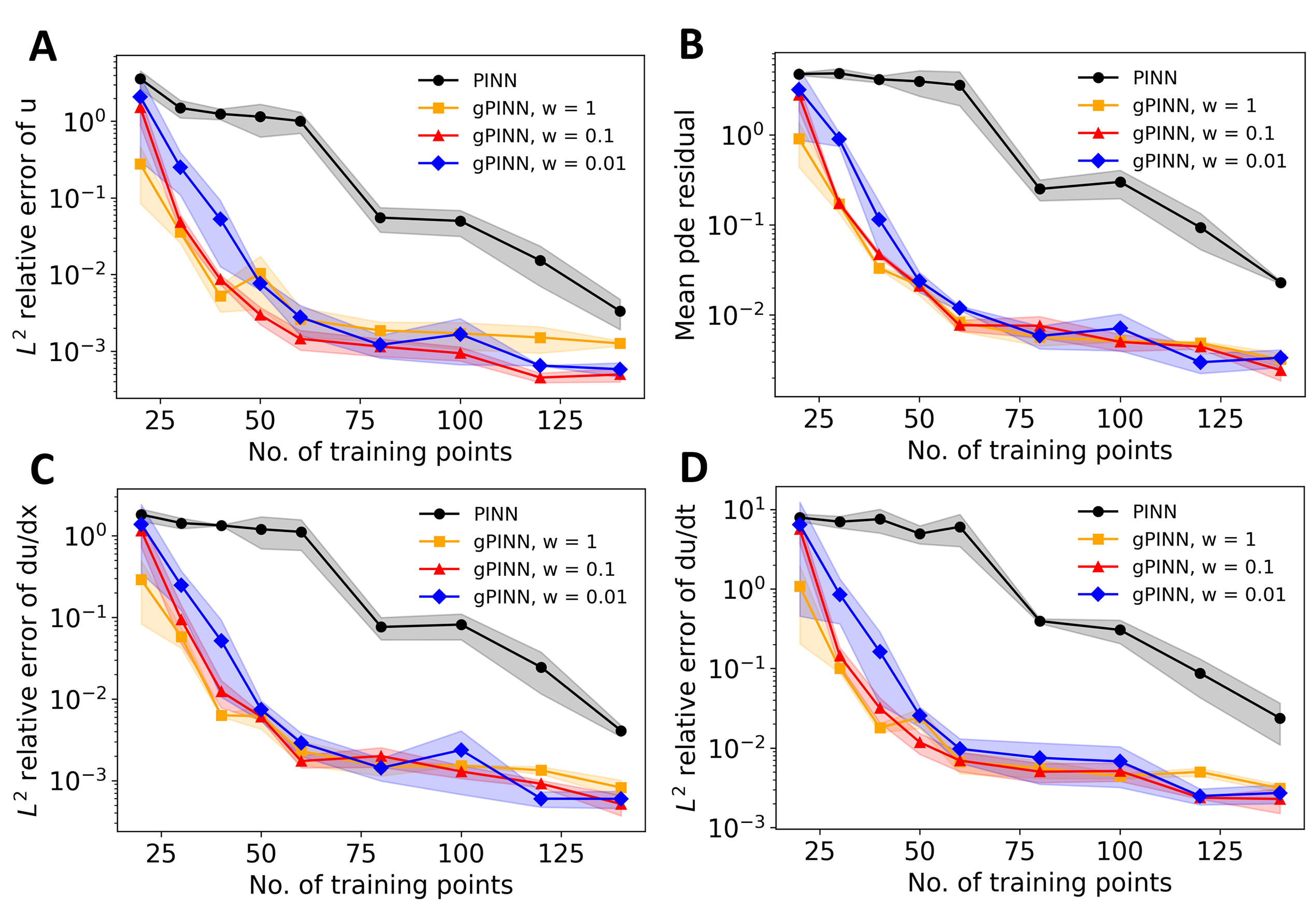}
    \caption{\textbf{Example in Section 3.2.2: Comparison between PINN and gPINN.} (\textbf{A}) $L^2$ relative error of $u$ for PINN and gPINN with $w=$1, 0.1, and 0.01. (\textbf{B}) Mean absolute value of the PDE residual. (\textbf{C}) $L^2$ relative error of $\frac{du}{dx}$. (\textbf{D}) $L^2$ relative error of $\frac{du}{dt}$.}
    \label{fig:3.2.2.1}
\end{figure}

In the Poisson equation, the performance of gPINN depends on the value of the weight, but in this diffusion-reaction system, gPINN is not sensitive to the value of $w$. gPINN with the values of $w=0.01$, 0.1, and 1 all outperform PINN by up to two orders of magnitude for the $L^2$ relative errors of $u$, $\frac{du}{dx}$ and $\frac{du}{dt}$, and the mean absolute error of the PDE residual (Fig.~\ref{fig:3.2.2.1}). gPINN reaches 1\% $L^2$ relative error of $u$ by using only 40 training points, while PINN requires more than 100 points to reach the same accuracy.

\begin{figure}[htbp]
    \centering
    \includegraphics[width=\textwidth]{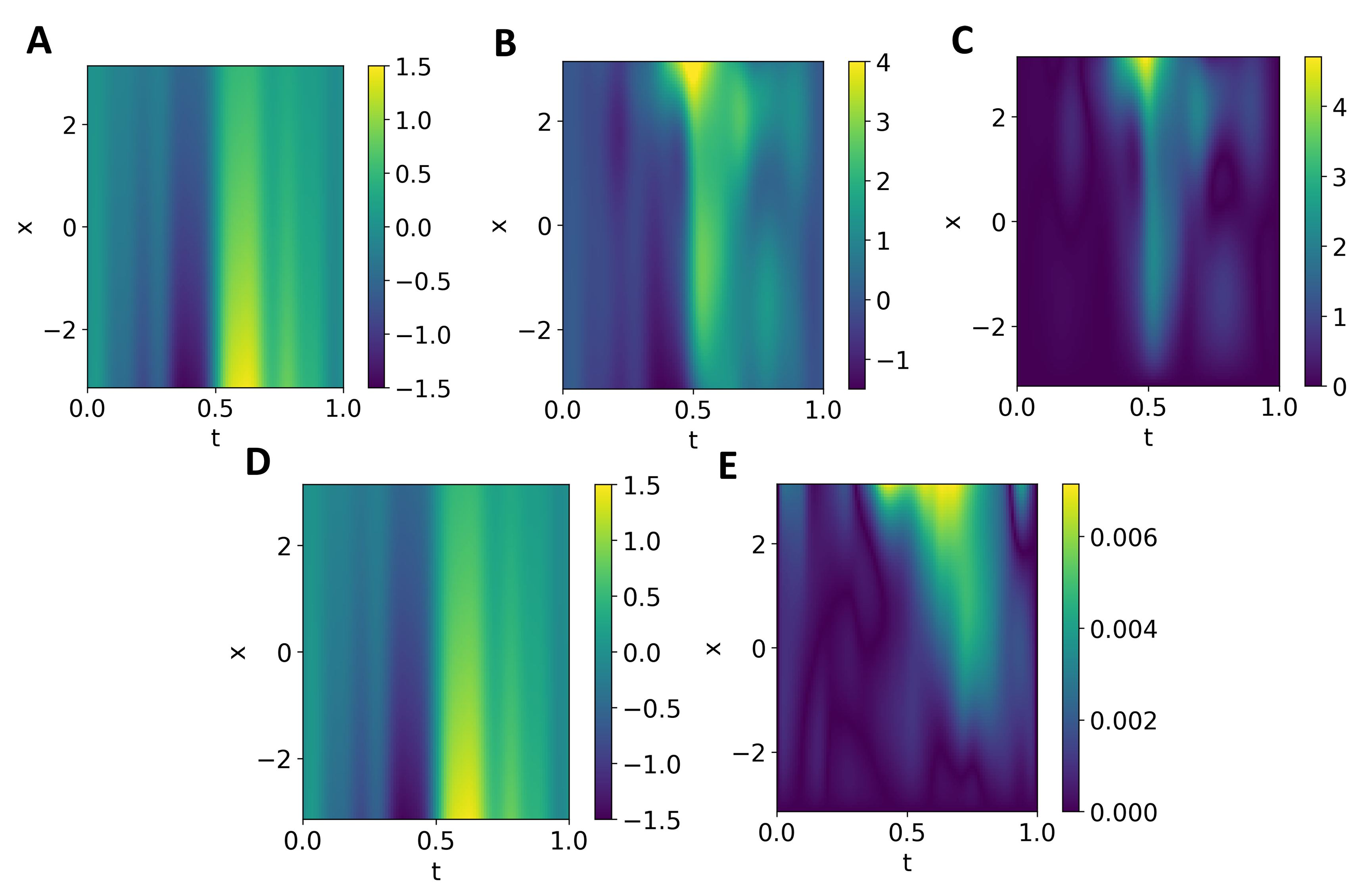}
    \caption{\textbf{Example in Section 3.2.2: Comparison between PINN and gPINN using 50 residual points for training.} (\textbf{A}) The exact solution in Eq.~\eqref{eq:dr_exact}. (\textbf{B} and \textbf{C}) The (B) prediction and (C) absolute error of PINN. (\textbf{D} and \textbf{E}) The (D) prediction and (E) absolute error of gPINN with $w=0.1$.}
    \label{fig:3.2.2.2}
\end{figure}

As an example, we show the exact solution, the predictions, and the error of PINN and gPINN with $w=0.01$ in Fig.~\ref{fig:3.2.2.2} when the number of the residual points is 50. The PINN prediction has a large error of about 100\%. However, the gPINN prediction's largest absolute error is around 0.007 and the $L^2$ relative error of 0.2\%.

\begin{figure}[htbp]
    \centering
    \includegraphics[width=\textwidth]{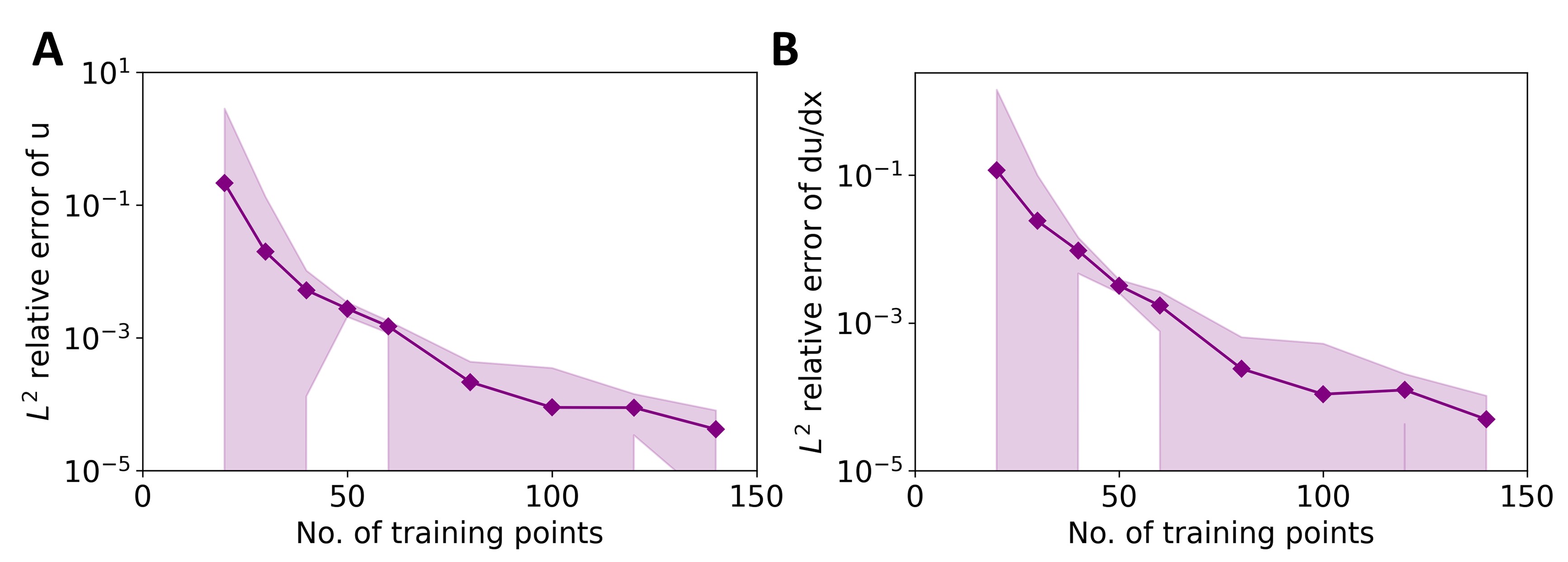}
    \caption{\textbf{Example in Section 3.2.2: gPINN is trained with a smaller learning rate and more iterations.} $L^2$ relative error of (\textbf{A}) $u$ and (\textbf{B}) $\frac{du}{dx}$ for gPINN with $w=$0.1.}
    \label{fig:3.2.2.3}
\end{figure}

We note that gPINN appears to plateau at around 130 training points (Fig.~\ref{fig:3.2.2.1}), which is due to network optimization. We show that a better accuracy is achieved using a smaller learning rate of $10^{-6}$ and more iterations ($5\times 10^6$) in Fig.~\ref{fig:3.2.2.3}. The $L^2$ relative error of $u$ and $\frac{du}{dx}$ decreases to less than 0.01\% using 140 training points and does not saturate.

\subsection{Inverse problems of PDEs}

In addition to solving forward PDE problems, we also apply gPINN for solving inverse PDE problems.

\subsubsection{Inferring the effective viscosity and permeability for the Brinkman-Forchheimer model}

The Brinkman-Forchheimer model can be viewed as an extended Darcy's law and is used to describe wall-bounded porous media flows:
\begin{equation*}
    -\frac{\nu_e}{\epsilon} \nabla^2 u + \frac{\nu u}{K} = g, \quad x\in [0, H],
\end{equation*}
where the solution $u$ is the fluid velocity, $g$ denotes the external force, $\nu$ is the kinetic viscosity of fluid, $\epsilon$ is the porosity of the porous medium, and $K$ is the permeability. The effective viscosity, $\nu_e$, is related to the pore structure and hardly to be determined. A no-slip boundary condition is imposed, i.e., $u(0) = u(1) = 0$. The analytic solution for this problem is 
\begin{equation*}
    u(x) = \frac{gK}{\nu} \left[1 - \frac{\cosh\left(r\left(x - \frac{H}{2} \right) \right)}{\cosh\left(\frac{r H}{2} \right)} \right]
\end{equation*}
with $r = \sqrt{\nu \epsilon / \nu_e  K}$. We choose $H=1$, $\nu_e = \nu = 10^{-3}$, $\epsilon=0.4$, and $K=10^{-3}$, and $g=1$. To infer $\nu_e$, we collect the data measurements of the velocity $u$ in only 5 sensor locations.

\begin{figure}[htbp]
    \centering
    \includegraphics[width=\textwidth]{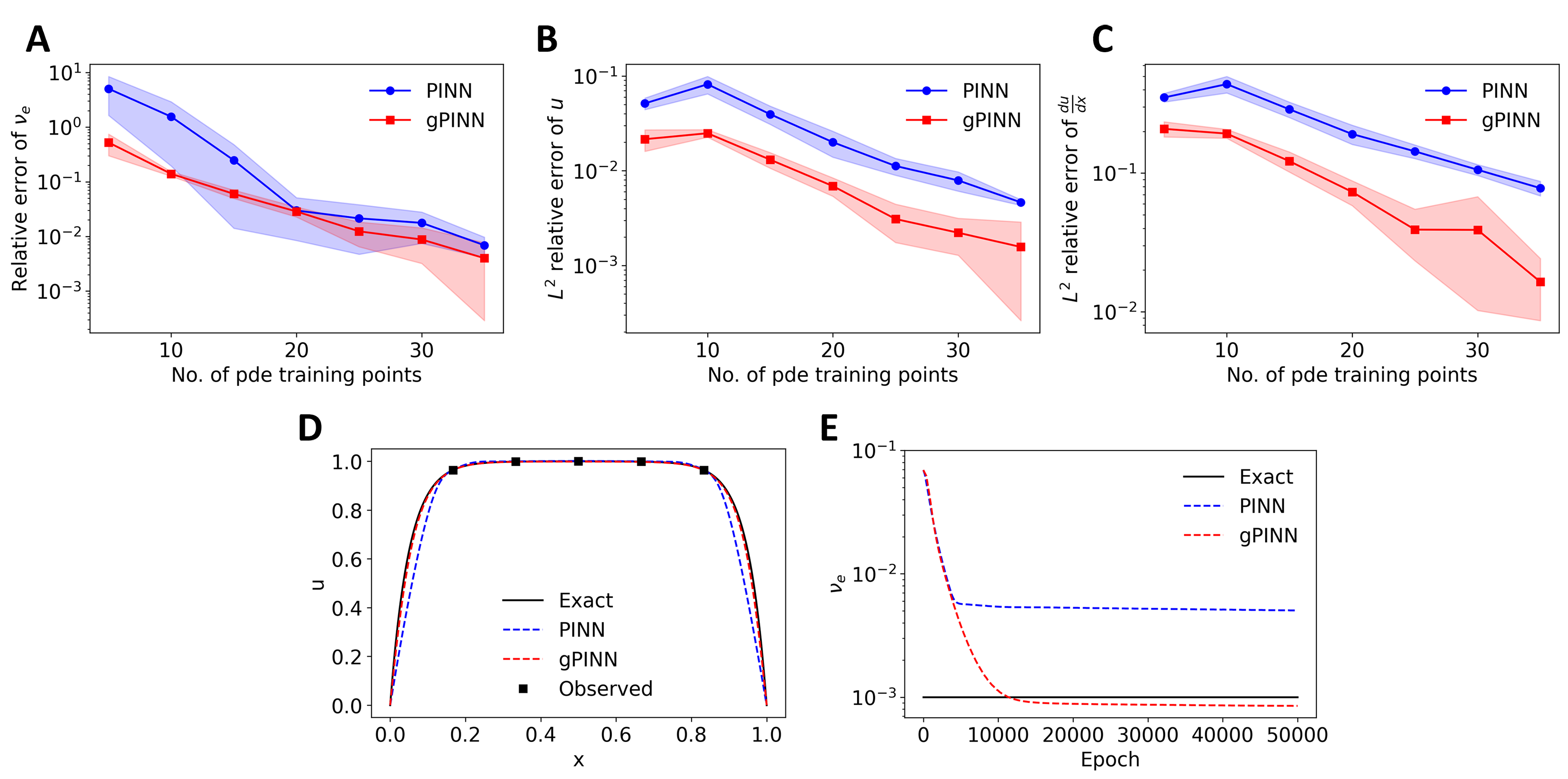}
    \caption{\textbf{Example in Section 3.3.1: Inferring $\nu_e$ by using PINN and gPINN from 5 measurements of $u$.} (\textbf{A}) Relative error of $\nu_e$. (\textbf{B}) $L^2$ relative error of $u$. (\textbf{C}) $L^2$ relative error of $\frac{du}{dx}$. (\textbf{D}) Example of the predicted $u$ using 5 observations of $u$ and 10 PDE residual points. The black squares in D show the observed locations. (\textbf{E}) The convergence of the predicted value for $\nu_e$ throughout training.}
    \label{fig:3.3.1.1}
\end{figure}

In PINN and gPINN, we simultaneously optimize the network and the unknown value of $\nu_e$. The loss weight in gPINN is chosen as $w = 0.1$. Similar as what we observed in the forward PDE problems, gPINN outperforms PINN in this case (Fig.~\ref{fig:3.3.1.1}). Specifically, the error of the predictions of gPINN for $u$ and $\frac{du}{dx}$ are about one order of magnitude smaller than PINN (Figs.~\ref{fig:3.3.1.1}B and C). Also, the inferred $\nu_e$ from gPINN is more accurate than that from PINN (Fig.~\ref{fig:3.3.1.1}A). We also show the example using only 10 PDE residual points. While PINN failed to predict $u$ near the boundary with a steep gradient, gPINN can still have a good accuracy (Fig.~\ref{fig:3.3.1.1}D). During the training, the predicted $\nu_e$ in PINN did not converge to the true value, while in gPINN the predicted $\nu_e$ is more accurate (Fig.~\ref{fig:3.3.1.1}E).

\begin{figure}[htbp]
    \centering
    \includegraphics[width=16cm]{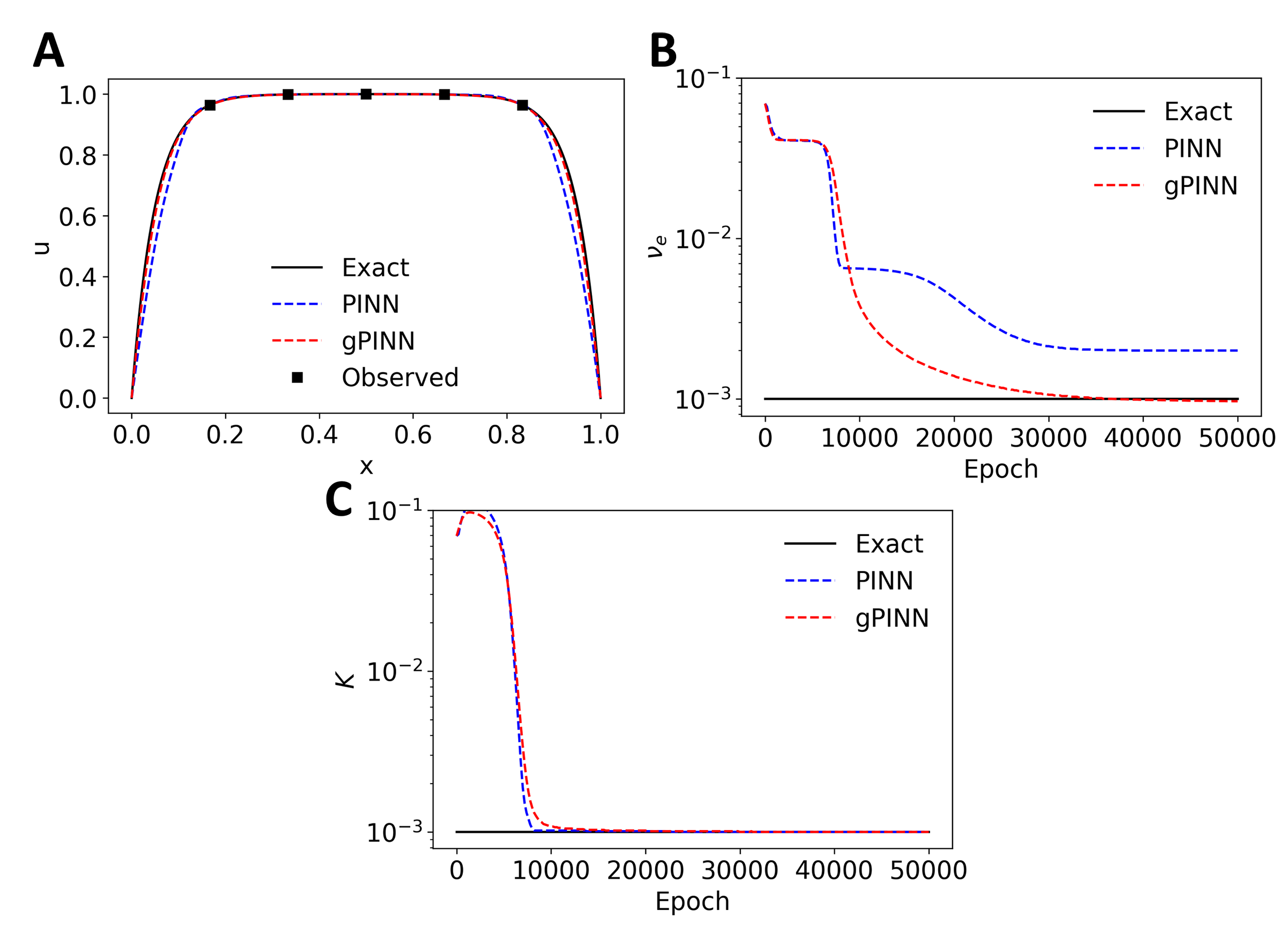}
    \caption{\textbf{Example in Section 3.3.1: Inferring both $\nu_e$ and $K$.} (\textbf{A}) The predicted $u$ from PINN and gPINN. (\textbf{B}) The convergence of the predicted value for $\nu_e$ throughout training. (\textbf{C}) The convergence of the predicted value for $K$. The black squares in A show the observed locations of $u$.}
    \label{fig:3.3.1.2}
\end{figure}

To further test the performance of gPINN, we next infer both $\nu_e$ and $K$ still from 5 measurements of $u$. Similarly, the PINN solution of $u$ struggles with the regions near the boundary, but gPINN can still achieve a good accuracy (Fig.~\ref{fig:3.3.1.2}A). Both PINN and gPINN converge to an accurate value of $K$ (Fig.~\ref{fig:3.3.1.2}C), but gPINN converges to much more accurate value for $\nu_e$ than PINN (Fig.~\ref{fig:3.3.1.2}B). 

\begin{figure}[htbp]
    \centering
    \includegraphics[width=16cm]{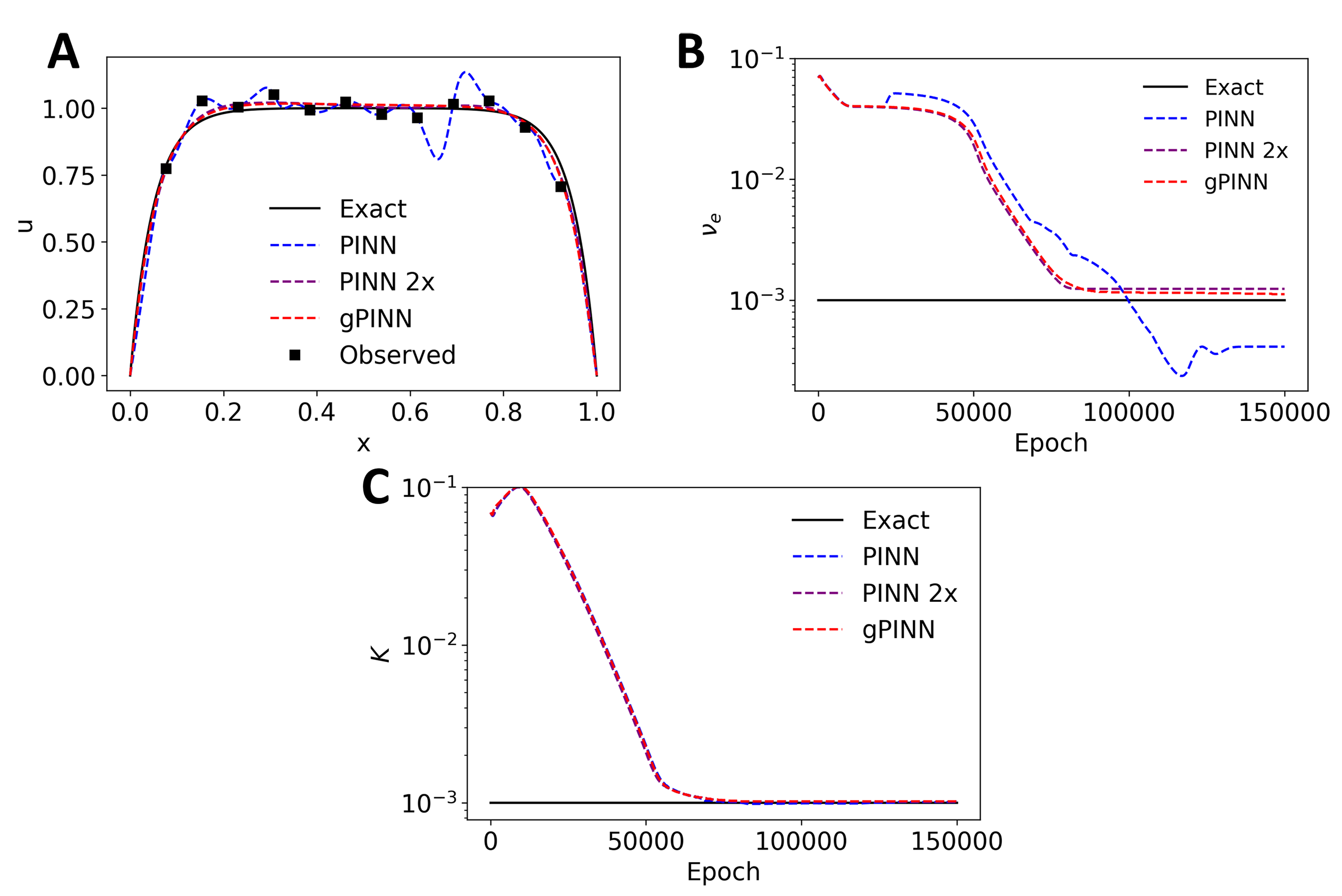}
    \caption{\textbf{Example in Section 3.3.1: Inferring both $\nu_e$ and $K$ from noise data.} (\textbf{A}) The predicted $u$ from PINN and gPINN. ``PINN 2x'' is PINN with twice more PDE training points. The black squares show the observed measurements of $u$. (\textbf{B}) The convergence of the predicted value for $\nu_e$ throughout training. (\textbf{C}) The convergence of the predicted value for $K$.}
    \label{fig:3.3.1.3}
\end{figure}

Next, we add Gaussian noise (mean 0 and standard deviation 0.05) to the observed values and infer both $\nu_e$ and $K$ using 12 measurements of $u$ (Fig.~\ref{fig:3.3.1.3}). Both PINN and gPINN converge to an accurate value for $K.$ Whereas PINN struggles with the added noise and struggles to learn $u$ and $\nu_e$, gPINN performs very well in both. However, after doubling the number of PDE training points from 15 to 30 (``PINN 2x'' in Fig.~\ref{fig:3.3.1.3}), PINN can also perform well at inferring $\nu_e$, though the performance is still slightly worse than that of gPINN.

\subsubsection{Inferring the space-dependent reaction rate in a diffusion-reaction system}

We consider a one-dimensional diffusion-reaction system in which the reaction rate $k(x)$ is a space-dependent function:
$$\lambda \frac{\partial^2 u}{\partial x^2} - k(x) u=f, \quad x \in[0,1],$$
where $\lambda = 0.01$ is the diffusion coefficient, $u$ is the solute concentration, and $f = \sin(2\pi x)$ is the source term. The objective is to infer $k(x)$ given measurements on $u$. The exact unknown reaction rate is
$$k(x) = 0.1 + \exp \left[-0.5 \frac{(x-0.5)^{2}}{0.15^{2}}\right].$$
In addition, the condition $u(x) = 0$ is imposed at $x = 0$ and 1.

\begin{figure}[htbp]
    \centering
    \includegraphics[width=15cm]{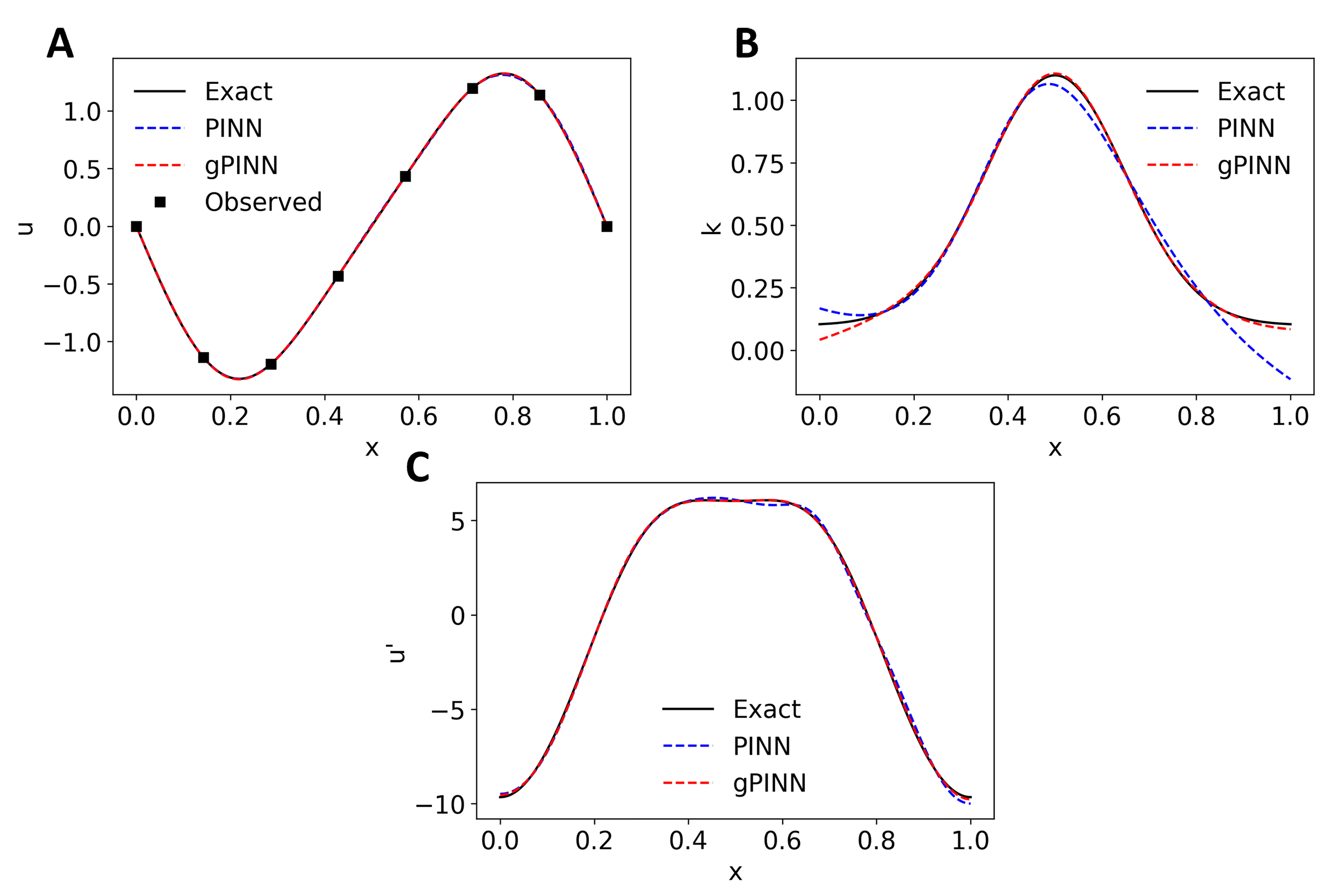}
    \caption{\textbf{Example in Section 3.3.2: Comparison between PINN and gPINN.} (\textbf{A}) The prediction of $u$. (\textbf{B}) The prediction of $k$. (\textbf{C}) The prediction of $\frac{du}{dx}$. We used 8 observations of $u$ (the black squares in A) and 10 residual points for training.}
    \label{fig:3.3.2}
\end{figure}

As the unknown parameter $k$ is a function of $x$ instead of just one constant, in addition to the network of $u$, we use another network to approximate $k$. We choose the weight $w=0.01$ in gPINN. We test the performance of PINN and gPINN by using 8 observations of $u$ and 10 PDE residual points for training. Both PINN and gPINN perform well in learning the solution $u$, though the PINN solution slightly deviates from the exact solution around $x = 0.8$ (Fig.~\ref{fig:3.3.2}A). However, for the inferred function $k$, gPINN's prediction was much more accurate than PINN (Fig.~\ref{fig:3.3.2}B). Also, the prediction of $\frac{du}{dx}$ by gPINN is more accurate than the prediction of PINN.

\subsection{gPINN enhanced by RAR}

To further improve the accuracy and training efficiency of gPINN for solving PDEs with a stiff solution, we apply RAR to adaptively improve the distribution of residual points during the training process.

\subsubsection{Burgers' equation}

We consider the 1D Burgers equation:
$$\frac{\partial u}{\partial t}+u \frac{\partial u}{\partial x}=\nu \frac{\partial^{2} u}{\partial x^{2}}, \quad x \in[-1,1], t \in[0,1],$$
with the initial and boundary conditions
$$u(x, 0)=-\sin (\pi x), \quad u(-1, t)=u(1, t)=0,$$
with $\nu = 0.01/\pi$.

\begin{figure}[htbp]
    \centering
    \includegraphics[width=10cm]{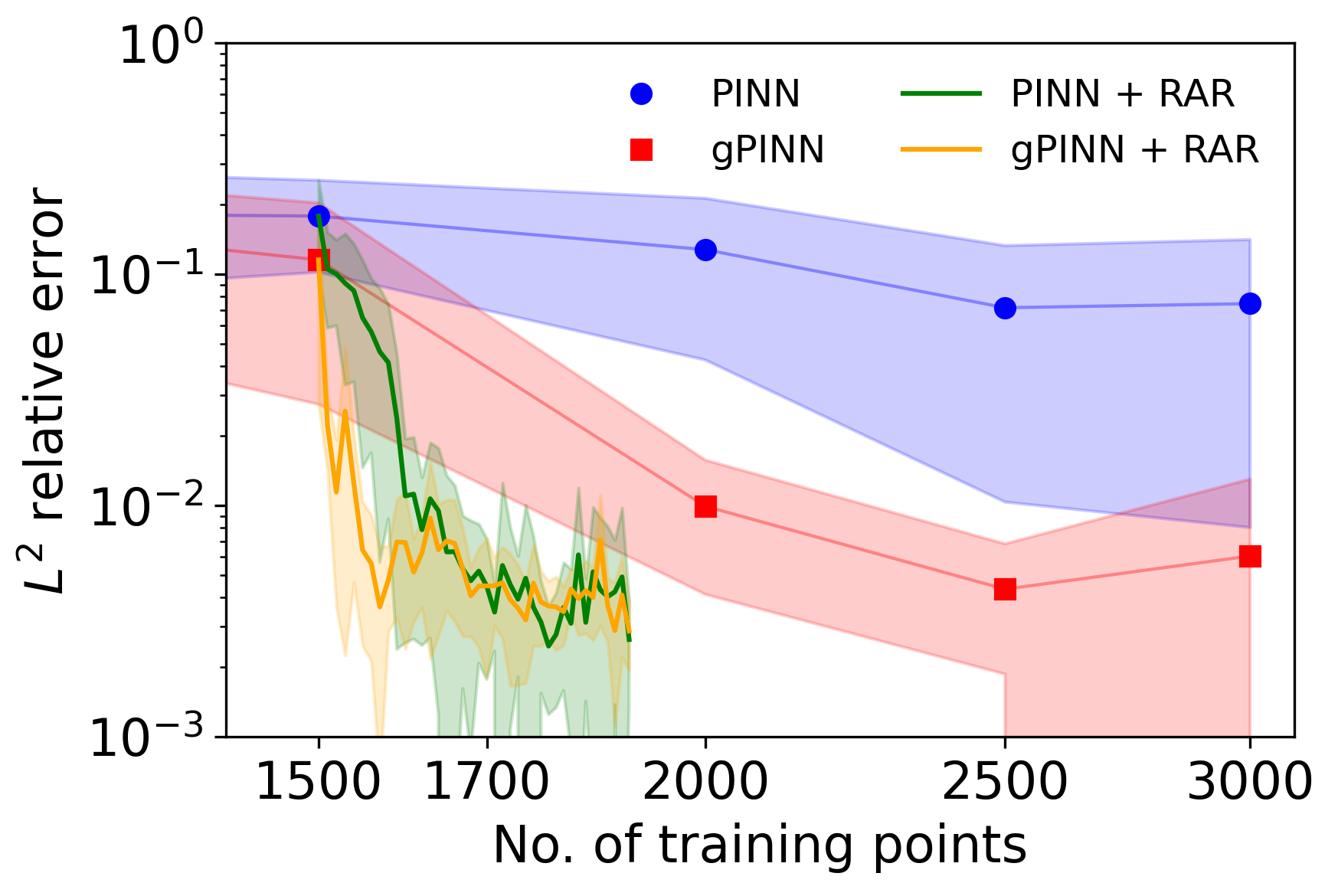}
    \caption{\textbf{Example in Section 3.4.1: $L^2$ relative errors of PINN, gPINN, PINN with RAR, and gPINN with RAR.} For RAR, we started from 1500 uniformly-distributed residual points and added 400 extra points.}
    \label{fig:3.4.1.3}
\end{figure}

We first test PINN and gPINN for this problem. PINN converges very slowly and has a large $L^2$ relative error ($\sim$10\%), while gPINN achieves one order of magnitude smaller error ($<$1\%) (the blue and red lines in Fig.~\ref{fig:3.4.1.3}), as we expected.

\begin{figure}[htbp]
\centering
\includegraphics[width=\textwidth]{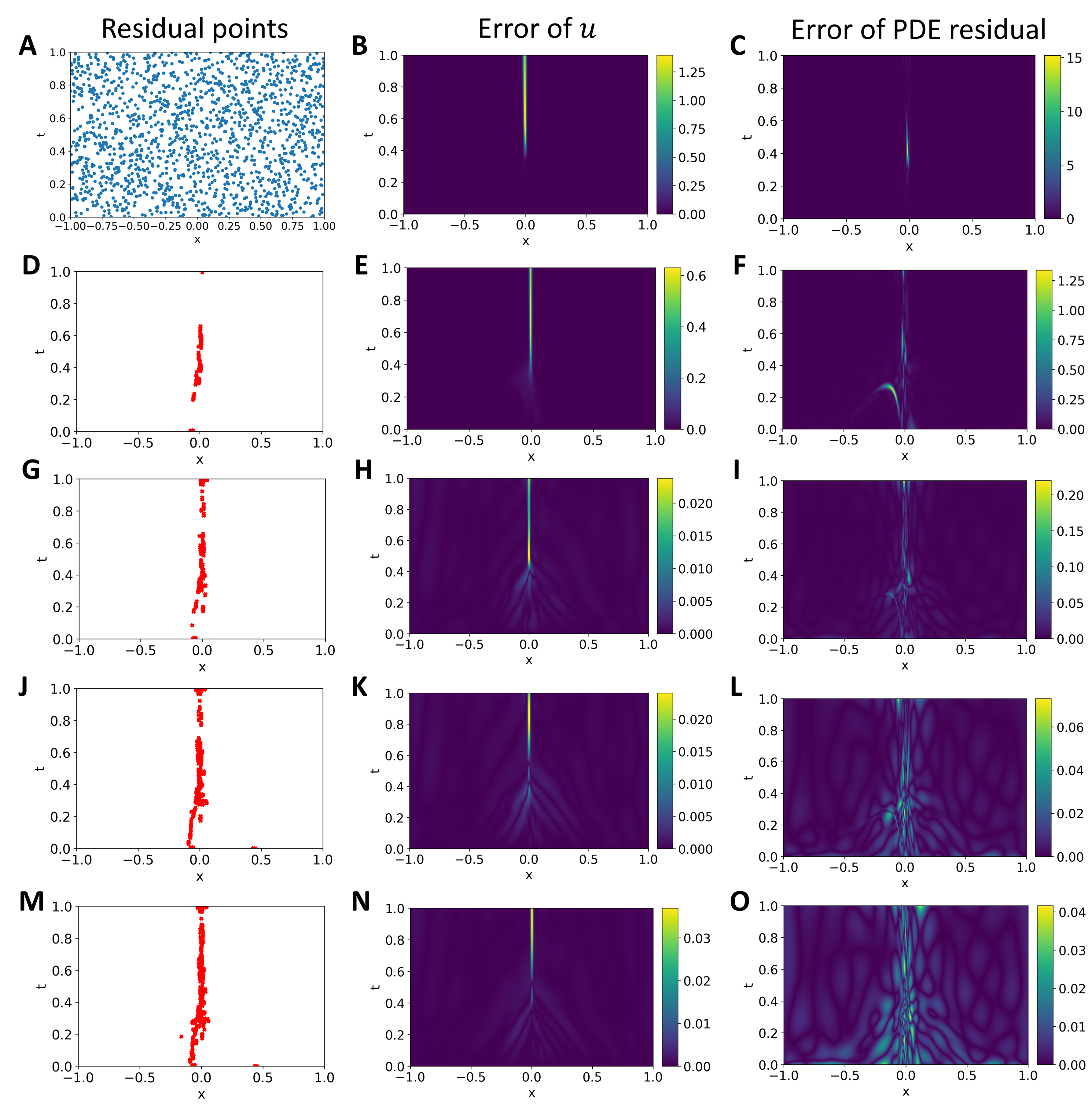}
\caption{\textbf{Example in Section 3.4.1: PINN with RAR.} (\textbf{A}, \textbf{B}, \textbf{C}) No extra points have been added. (A) The initial distribution of the 1500 residual points. (B) The absolute error of $u$. (C) The absolute error of the PDE residual. (\textbf{D}, \textbf{E}, \textbf{F}) 100 extra points (point locations shown in D) have been added. (\textbf{G}, \textbf{H}, \textbf{I}) 200 extra points (point locations shown in G) have been added. (\textbf{J}, \textbf{K}, \textbf{L}) 300 extra points (point locations shown in J) have been added. (\textbf{M}, \textbf{N}, \textbf{O}) 400 extra points (point locations shown in M) have been added.}
\label{fig:3.4.1.1}
\end{figure}

The solution to this 1D Burgers' equation is very steep near $x=0$, so intuitively there should be more residual points around that region. We first show the effectiveness of PINN with RAR proposed in Ref.~\cite{lu2021deepxde}. For RAR, we first train the network using 1500 uniformly-distributed residual points and then gradually add 400 more residual points during training. We added 10 new points at a time, i.e., $m=10$ in Algorithm~\ref{alg:rar}. For the Burgers' equation, the solution has a steep gradient around $x=0$, and after the initial training of 1500 residual points, the region around $x=0$ has the largest error of $u$ (Fig.~\ref{fig:3.4.1.1}B) and the PDE residual (Fig.~\ref{fig:3.4.1.1}C). RAR automatically added new points near the largest error, as shown in the left column in Fig.~\ref{fig:3.4.1.1}, and then the errors of $u$ and the PDE residual consistently decreases as more points are added (Fig.~\ref{fig:3.4.1.1}). By using RAR, the error of PINN decreases very fast, and PINN achieves the $L^2$ relative error of $\sim$0.3\% by using only 1900 residual points for training (the green line in Fig.~\ref{fig:3.4.1.3}), which is even better than gPINN with more training points.

\begin{figure}[htbp]
\centering
\includegraphics[width=\textwidth]{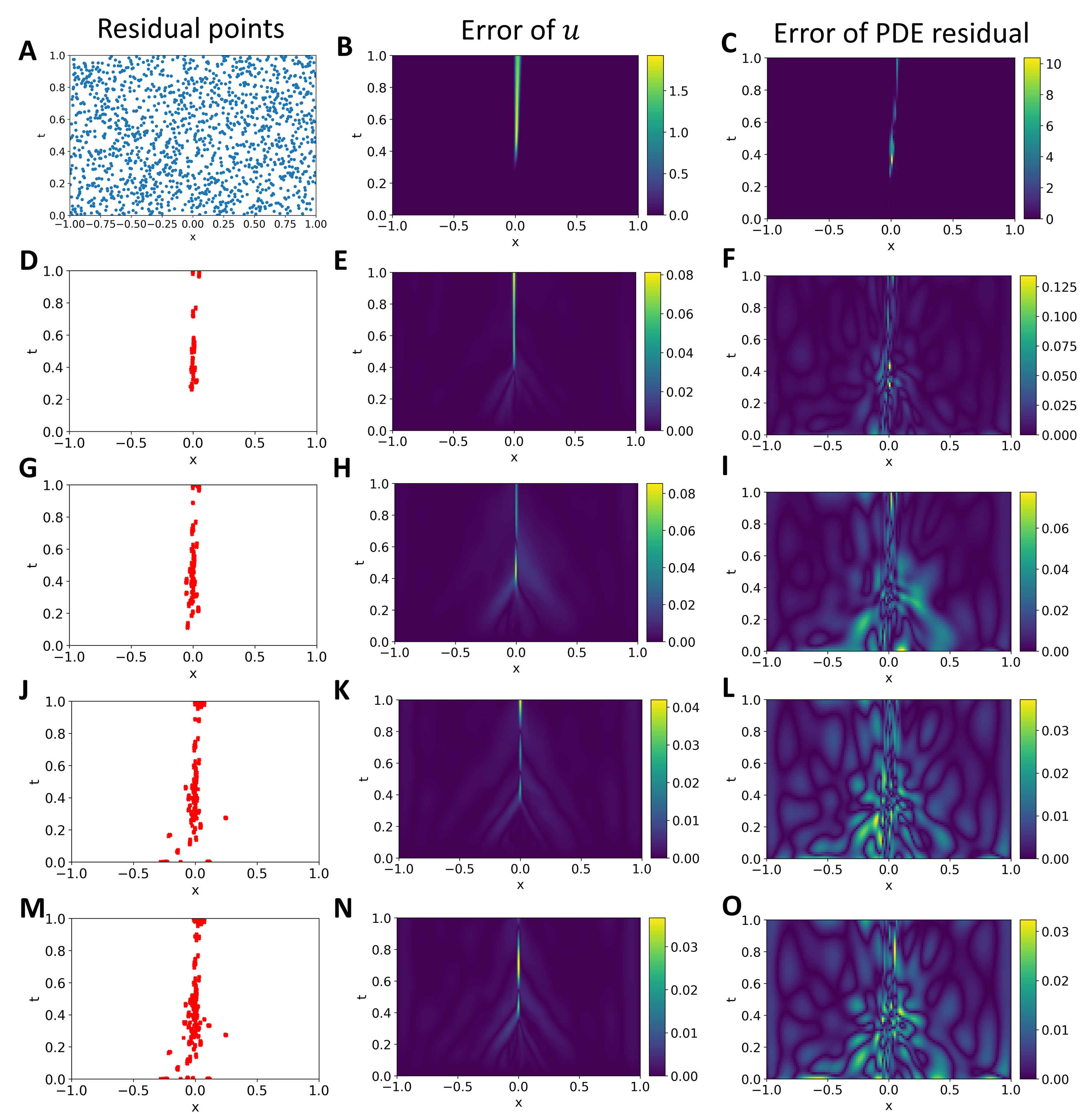}
\caption{\textbf{Example in Section 3.4.1: gPINN with RAR.} (\textbf{A}, \textbf{B}, \textbf{C}) No extra points have been added. (A) The initial distribution of the 1500 residual points. (B) The absolute error of $u$. (C) The absolute error of the PDE residual. (\textbf{D}, \textbf{E}, \textbf{F}) 100 extra points (point locations shown in D) have been added. (\textbf{G}, \textbf{H}, \textbf{I}) 200 extra points (point locations shown in G) have been added. (\textbf{J}, \textbf{K}, \textbf{L}) 300 extra points (point locations shown in J) have been added. (\textbf{M}, \textbf{N}, \textbf{O}) 400 extra points (point locations shown in M) have been added.}
\label{fig:3.4.1.2}
\end{figure}

We also used gPINN together with RAR. gPINN with RAR also added new points near $x=0$, and the errors of $u$ and the PDE residual consistently decrease when more points are added adaptively (Fig.~\ref{fig:3.4.1.2}), similarly to PINN with RAR.
The final accuracy of PINN with RAR and gPINN with RAR is similar, but the error of gPINN with RAR drops much faster than PINN with RAR when using only 100 extra training points. Therefore, by pairing gPINN together with the RAR, we can achieve the best performance.

\subsubsection{Allen--Cahn equation}

We also consider the following Allen--Cahn equation:
$$\frac{\partial u}{\partial t} = D \frac{\partial^2 u}{\partial x^2} + 5(u - u^3), \quad x \in[-1, 1], ~ t \in [0, 1],$$
with the initial and boundary conditions:
\begin{gather*}
    u(x, 0) = x^2 \cos(\pi x), \\
    u(-1, t) = u(1, t) = -1.
\end{gather*}
where $D = 0.001$. The solution to this Allen--Cahn equation has multiple very steep regions similar to that of the Burgers equation.

\begin{figure}[htbp]
    \centering
    \includegraphics[width=10cm]{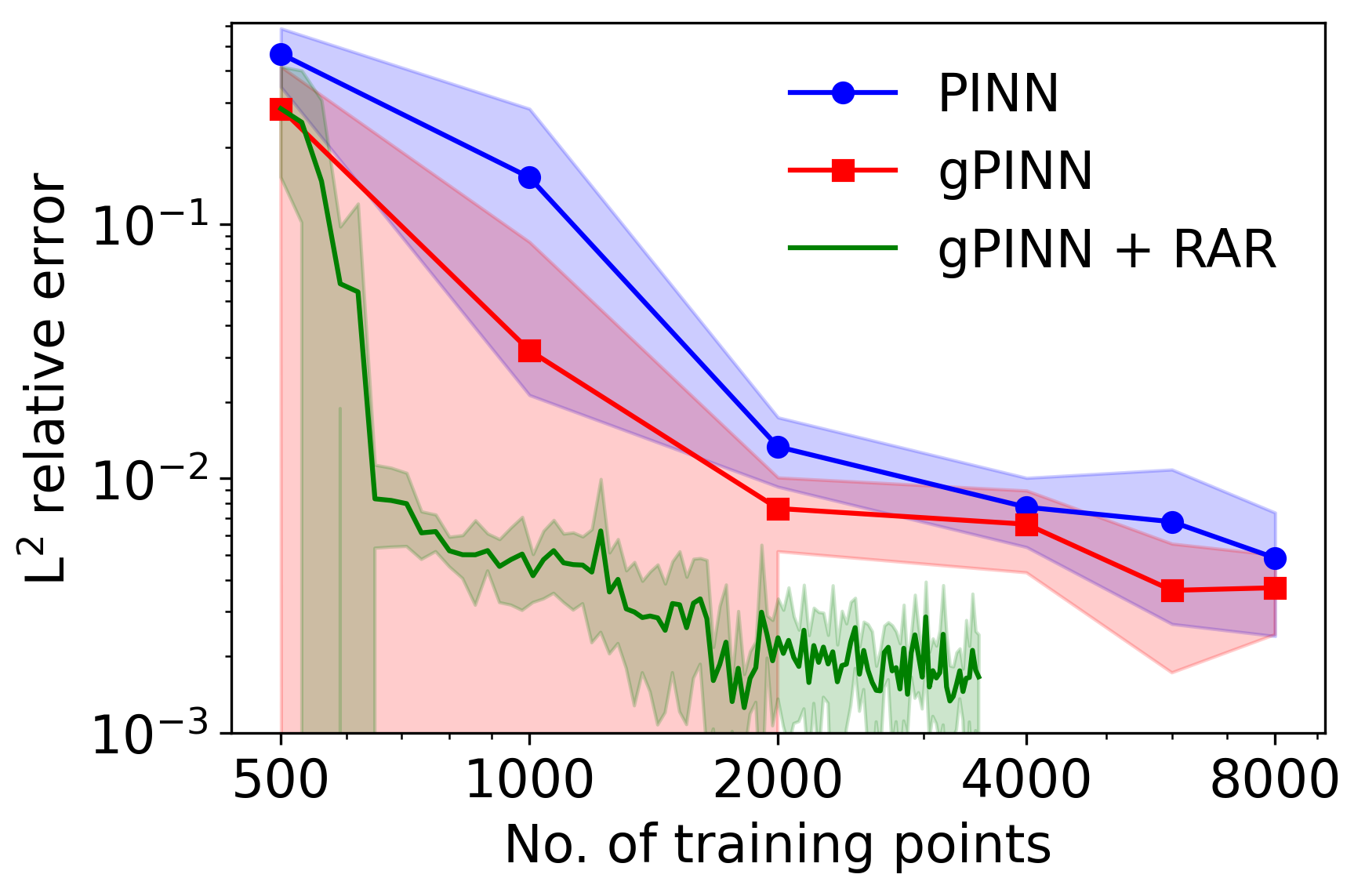}
    \caption{\textbf{Example in Section 3.4.2: $L^2$ relative of PINN, gPINN, and gPINN with RAR.} For RAR, there were 500 initial points and 3000 added points.}
    \label{fig:3.4.2.2}
\end{figure}

First, comparing PINN and gPINN, we can once again observe that gPINN has better accuracy than PINN (the blue and red lines in Fig.~\ref{fig:3.4.2.2}). gPINN requires around 2000 training points to reach $1\%$ error, while PINN requires around 4000 training points to reach that same accuracy.

\begin{figure}[htbp]
\centering
\includegraphics[width=\textwidth]{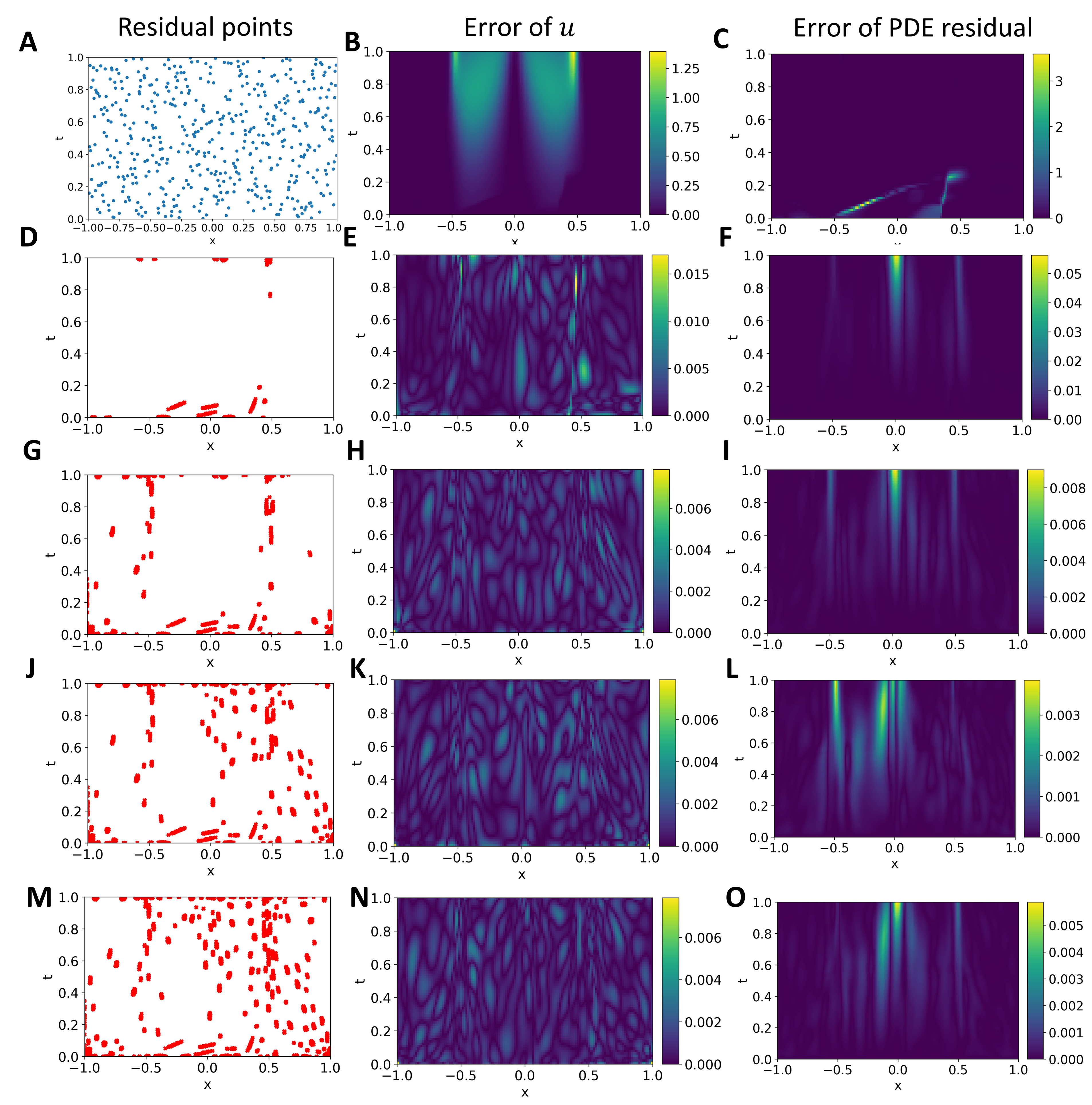}
\caption{\textbf{Example in Section 3.4.2: gPINN with RAR.} (\textbf{A}, \textbf{B}, \textbf{C}) No extra points have been added. (A) The initial distribution of the 500 residual points. (B) The absolute error of $u$. (C) The absolute error of the PDE residual. (\textbf{D}, \textbf{E}, \textbf{F}) 300 extra points (point locations shown in D) have been added. (\textbf{G}, \textbf{H}, \textbf{I}) 1200 extra points (point locations shown in G) have been added. (\textbf{J}, \textbf{K}, \textbf{L}) 2100 extra points (point locations shown in J) have been added. (\textbf{M}, \textbf{N}, \textbf{O}) 3000 extra points (point locations shown in M) have been added.}
\label{fig:3.4.2.1}
\end{figure}

We next show the behavior and effectiveness of gPINN with RAR again. We first train the network using 500 uniformly-distributed residual points and then gradually add 3000 more residual points during training with 30 training points added at a time. The solution $u$ has two peaks around $x=-0.5$ and $x=0.5$, where the largest error occurs (Fig.~\ref{fig:3.4.2.1}B). The added points by RAR also fall on these two regions of high error, as shown in Fig.~\ref{fig:3.4.2.1}G. The error becomes nearly uniform with 1200 added points (Fig.~\ref{fig:3.4.2.1}H), and then the added points start to become more and more uniform (Figs.~\ref{fig:3.4.2.1}J and M). By using RAR, the error of gPINN decreases drastically fast, and by adding only 200 additional points (i.e., 700 in total), gPINN reaches $1\%$ error. However, gPINN with RAR begins to plateau at about 1500 training points with approximately $0.1\%$ error, which could be resolved by using a smaller learning rate as we show in Section~\ref{sec:diffusion-reaction}.

\section{Conclusion}
\label{sec:conclusion}

In this paper, we proposed a new version of physics-informed neural networks (PINNs) with gradient enhancement (gPINNs) for improved accuracy and training efficiency. We demonstrated the effectiveness of gPINN in both forward and inverse PDE problems, including Poisson equation, diffusion-reaction equation, Brinkman-Forchheimer model, Burgers' equation, and Allen-Cahn equation. Our numerical results from all of the examples show that gPINN clearly outperforms PINN with the same number of training points in terms of the $L^2$ relative errors of the solution and the derivatives of the solution. For the inverse problems, gPINN learned the unknown parameters more accurately than PINN. In addition, we combined gPINN with residual-based adaptive refinement (RAR) to further improve the performance. For the PDEs with solutions that had especially steep gradients, such as Burgers' equation and the Allen-Cahn equation, RAR allowed gPINN to perform well with much fewer residual points.

When using the same number of residual points, gPINN achieves better accuracy than PINN, but the computational cost of gPINN is higher than PINN because of the additional loss terms with higher order derivatives. In our examples, the cost of gPINN relative to PINN was typically 2 to 3 times greater. In some cases, the performance of PINN with twice more training points (i.e., similar computational cost as gPINN) is similar to gPINN, but in some cases (e.g., the Burgers' equation), gPINN still performs better than PINN even when PINN uses twice more points. In addition, in this work we computed higher-order derivatives by applying automatic differentiation (AD) of first-order derivative recursively. However, this nested approach is not efficient enough~\cite{baydin2018automatic,margossian2019review}, and other methods, e.g., Taylor-mode AD, have been developed for better computational performance~\cite{bettencourt2019taylor}.

Compared to PINN, in gPINN we have an extra hyperparameter---the weight coefficient of the gradient loss. In some problems, the performance of gPINN is not sensitive to this weight, but in some cases, there exists an optimal weight to achieve the best accuracy, and thus we need to tune the weight. Considering that there are already many hyperparameters such as network depth/width, learning rate, and training epochs, only one extra hyperparameter is not a worrying issue. However, it is still an interesting research topic in the future to automatically determine an optimal weight. Moreover, it is possible to combine gPINN with other extensions of PINN to further improve the performance, such as extended PINN (XPINN)~\cite{jagtap2020extended} and parareal PINN (PPINN)~\cite{meng2020ppinn}.

\section*{Acknowledgements}

This work was supported by the DOE PhILMs project (no. DE-SC0019453) and OSD/AFOSR MURI grant FA9550-20-1-0358. J.Y. and L.L. thank MIT's PRIMES-USA program.

\bibliography{main}

\end{document}